\pdfoutput=1

\documentclass[journal]{IEEEtran}

\usepackage{cite}
\ifCLASSINFOpdf
  \usepackage[pdftex]{graphicx}
  \graphicspath{{../pdf/}{../jpeg/}}
   \DeclareGraphicsExtensions{.pdf,.jpeg,.png}
\else
   \usepackage[dvips]{graphicx}
   \graphicspath{{../eps/}}
   \DeclareGraphicsExtensions{.eps}
\fi
\ifCLASSOPTIONcompsoc
\usepackage[caption=false,font=normalsize,labelfon
t=sf,textfont=sf]{subfig}
\else
\usepackage[caption=false,font=footnotesize]{subfi
	g}
\fi

\usepackage{amsmath}
\usepackage{amsfonts} 
\usepackage{algorithm}
\usepackage{algorithmic}

\usepackage{booktabs} 

\begin{document}

\title{Second-order Approximation of Minimum Discrimination Information in Independent Component Analysis}

\author{YunPeng ~Li%\quad ZhaoHui ~Ye
	
	% <-this % stops a space
	\thanks{YunPeng Li was with the Department of Automation, Tsinghua University, Beijing, China e-mail: liyp18@tsinghua.org.cn}
%	\thanks{ZhaoHui Ye is with the Department of Automation, Tsinghua University, Beijing, China e-mail: yezhaohui@mail.tsinghua.edu.cn}
}


\markboth{Journal of \LaTeX\ Class Files,~Vol.~14, No.~8, August~2015}%
{Shell \MakeLowercase{\textit{et al.}}: Bare Demo of IEEEtran.cls for IEEE Journals}



\maketitle

\begin{abstract}
 Independent Component Analysis (ICA) is intended to recover the mutually independent sources from their linear mixtures, and $FastICA$ is one of the most successful ICA algorithms. Although it seems reasonable to improve the performance of $FastICA$ by introducing more nonlinear functions to the negentropy estimation, the original fixed-point method (approximate Newton method) in $FastICA$ degenerates under this circumstance. To alleviate this problem, we propose a novel method based on the second-order approximation of minimum discrimination information (MDI). The joint maximization in our method is consisted of minimizing single weighted least squares and seeking unmixing matrix by the fixed-point method. Experimental results validate its efficiency compared with other popular ICA algorithms.

\end{abstract}


% \fi
\IEEEpeerreviewmaketitle

\section{Introduction}
\IEEEPARstart{I}{ndependent} Component Analysis (ICA) has been widely used in diverse fields, such as machine learning, signal processing, and stats. Given the $m$ dimensional observed mixtures $\mathbf{x}=(x_{1},\cdots,x_{m})^{T}$, independent component analysis models them as the linear combination of $m$ independent sources $\mathbf{s}=(s_{1},\cdots,s_{m})^{T}$, 
\begin{equation}
\label{eq:mix}
\mathbf{x} = \mathbf{A}\mathbf{s}
\end{equation}
where the mixing matrix $\mathbf{A}\in \mathbb{R}^{m\times m}$. Given $N$ independent identically distributed samples of $\mathbf{x}$, the goal of ICA is to recover the unknown sources $\mathbf{s}$ and estimate the mixing matrix $\mathbf{A}$. We model the recovering process as,
\begin{equation}
\label{eq:unmix}
\mathbf{y} = \mathbf{W}\mathbf{x}
\end{equation}
where the sources' estimation $\mathbf{y}=(y_{1},\cdots,y_{m})^{T}$ is the scaling and permutation of sources $\mathbf{s}$, and $\mathbf{W} \in \mathbb{R}^{m\times m}$ is called the unmixing matrix. In general, one assumes that $\mathrm{E}(\mathbf{s})=\mathbf{0}$ and $\mathrm{Cov}(\mathbf{s})=\mathbf{I}$. It has been shown that $\mathbf{W}$ is identifiable up to scaling and permutation of its rows if at most one $s_{i}$ is Gaussian \cite{Comon1994}. Since there often exists centering and whitening preprocessing stages for the observation $\mathbf{x}$, $\mathbf{W}$ is restricted to be an orthonormal matrix $\mathbf{W}\mathbf{W}^{T}=\mathbf{I}$.

Many approaches have been proposed for ICA in the past researches\cite{Aapo2001,Pierre2010}, and the most two popular types of ICA algorithms seem to be the maximum likelihood estimation and the contrast function approaches. Parametric maximum likelihood estimation (density matching)\cite{Pham92,BellSej1995,mky1996,JC1997} is used to infer the parametric model in ICA by specifying the distributions for the component $s_{i}$. Unfortunately, the performances of these parametric methods are highly dependent on the prior assumptions on the unknown sources. To keep the components of $\mathbf{s}$ unspecified, several nonparametric ICA\cite{HaTi2002,Alexander2004,chen2006,Sam2012} are proposed at the cost of high computation burden or the difficult selection of tuning parameters. In the contrast function approaches, several criteria are chosen to represent the measure of independence or non-Gaussianity, for example, the mutual information \cite{Comon1994,BellSej1995}, the nonlinear decorrelation \cite{JUTTEN19911,Bach2003}, higher-order moments\cite{JCar1989,JCar1993}, and the entropy \cite{fastica1999,RADICAL,Kol2006}.  For other recent approaches, see also\cite{Ilmonen2011,Matteson2017,Pierre2018,spurek2018,Anastasia2019}.

$FastICA$\cite{fastica1999} is one of the most successful ICA algorithms, whose contrast function (approximation of negentropy\cite{Hyv1998}) is defined as the expectation of a single nonlinear function. $FastICA$ enjoys low computation and fast convergence due to the efficient fixed-point method\cite{fastica1999}, which is equivalent to the approximate Newton method without calculating the inverse of the Hessian matrix. Unfortunately, $FastICA$ usually fails when there is a great mismatching between its single nonlinear function and the unknown sources' distributions. Although we can introduce more nonlinear functions to improve the negentropy estimation (suggested by research \cite{Hyv1998}), the original fixed-point method in $FastICA$ fails (Hessian matrix can not be approximated to a diagonal matrix).

In this paper, we present a novel ICA algorithm $MDIICA$ based on the second-order approximation of minimum discrimination information (MDI)\cite{gokhale1978}. Although conceptually, our work seems similar to the approximation of negentropy used in $FastICA$,  they are quite different in derivations. In addition, the fixed-point method can be directly applied to our method at the negligible cost of computation, when more nonlinear functions are required to improve the negentropy estimation.
In Section \ref{section:2}, we explain the difficulties in $FastICA$, when its contrast function is composed of several nonlinear functions. Section \ref{section:3} presents the derivations of our novel ICA algorithm. In Section \ref{section:4}, we compare our method with other known algorithms in both simulation and real data experiment. We conclude our contributions in Section \ref{section:5}.

\section{Difficulties in $FastICA$}
\label{section:2}
In this section, we firstly review the contrast function (expectation of a single nonlinear function) and the fixed-point method used in $FastICA$, then we reveal the difficulties when more nonlinear functions are used.

To separate the sources from their mixtures, $FastICA$ maximizes the approximation of negentropy\cite{Hyv1998} $J(\mathbf{w})$
\begin{equation}
\label{eq:neg}
\begin{aligned}
\max_{\mathbf{w}} \quad & J(\mathbf{w}) = \frac{1}{2}\sum_{k=1}^{p} \mathrm{E}\{G_{k}(\mathbf{w}^{T}\mathbf{x})\}^{2}\,
\quad\mathrm{s.t.}  &  \mathbf{w}^{T}\mathbf{w}=1
\end{aligned}
\end{equation}
where $\mathbf{w}$ is the row of the unmixing matrix $\mathbf{W}$, $\{G_{k}(.)\}_{k=1}^{p}$ are $p$ nonlinear functions. These nonlinear functions should satisfy the following constraints\cite{Hyv1998},
\begin{equation}
\label{eq:constraints1}
\int\phi(x)G_{i}(x)G_{j}(x)dx =   
  \begin{cases}
1\quad\mbox{$i=j$}\\
0\quad\mbox{$i\ne j$}
\end{cases}
\end{equation}
\begin{equation}
\label{eq:constraints2}
\int \phi(x)G_{i}(x)x^{k}dx = 0 \quad k=0,1,2
\end{equation}
where $\phi(.)$ is the standard Gaussian function. In practice, one can take any set of linearly independent  functions $\{\overline{G}_{k}(.)\}_{k=1}^{p}$, and apply Gram-Schmidt orthonormalization on the selected set to satisfy the assumptions above\cite{Hyv1998}. Fortunately, such computation can be simplified if a single nonlinear function is used. When $p=1$ in $FastICA$, the maximal of $J(\mathbf{w})=\frac{1}{2}\mathrm{E}\{G_{1}(\mathbf{w}^{T}\mathbf{x})\}^{2}$ are obtained at certain optima of $\mathrm{E}\{G_{1}(\mathbf{w}^{T}\mathbf{x})\}$, the equivalent optimization problem in $FastICA$ becomes,
 \begin{equation}
 \label{eq:neg2}
 \begin{aligned}
 \max_{\mathbf{w,\lambda}} /\min_{\mathbf{w,\lambda}}\quad & \mathcal{L}(\mathbf{w},\lambda)=\mathrm{E}\{G_{1}(\mathbf{w}^{T}\mathbf{x})\} -\lambda (\mathbf{w}^{T}\mathbf{w}-1) \\
 \end{aligned}
 \end{equation}
where $\lambda$ is the Lagrange multiplier, and the gradient $\nabla\mathcal{L}(\mathbf{w},\lambda)$ is
\begin{equation}
\label{eq:grad}
\nabla\mathcal{L}(\mathbf{w},\lambda) = \mathrm{E}\{G^{'}_{1}(\mathbf{w}^{T}\mathbf{x})\mathbf{x}\}-2\lambda\mathbf{w} =\mathbf{0}
\end{equation}
It is easy to find that $\lambda=\frac{1}{2}\mathrm{E}\{G^{'}_{1}(\mathbf{w}^{T}\mathbf{x})\mathbf{w}^{T}\mathbf{x}\}$, and the Hessian matrix $\nabla^{2}\mathcal{L}(\mathbf{w},\lambda)$ is 
\begin{equation}
\label{eq:hessian}
\begin{aligned}
\nabla^{2}\mathcal{L}(\mathbf{w},\lambda) &= \mathrm{E}\{G^{''}_{1}(\mathbf{w}^{T}\mathbf{x})\mathbf{x}\mathbf{x}^{T}\}-2\lambda\mathbf{I}\\
&\approx \mathrm{E}\{G^{''}_{1}(\mathbf{w}^{T}\mathbf{x})\}\mathrm{E}\{\mathbf{x}\mathbf{x}^{T}\}-2\lambda\mathbf{I}\\
&\approx \mathrm{E}\{G^{''}_{1}(\mathbf{w}^{T}\mathbf{x})\}\mathbf{I}-2\lambda\mathbf{I}\\
\end{aligned}
\end{equation}
Thanks to the whitening stage ($\mathrm{E}\{\mathbf{x}\mathbf{x}^{T}\}=\mathbf{I}$) on the observation $\mathbf{x}$ , the approximation in equation (\ref{eq:hessian}) is available. Thus, the Hessian matrix is diagonal and easy to be inverted, and the fixed-point method (approximate Newton method) can be directly applied in $FastICA$.

The concise of $FastICA$ is due in large part to the selection of a single nonlinear function. Unfortunately, these advantages are diminished, when more nonlinear functions are required. To use more nonlinear functions, the Gram-Schmidt orthonormalization in equation (\ref{eq:constraints1}) can not be omitted anymore, and the original fixed-point method fails in $FastICA$. The equivalent optimization problem concerning equation (\ref{eq:neg}) develops into the following form
 \begin{equation}
\label{eq:neg3}
\begin{aligned}
\max_{\mathbf{w,\lambda}} \quad & \mathcal{L}(\mathbf{w},\lambda)=\frac{1}{2}\sum_{k=1}^{p} \mathrm{E}\{G_{k}(\mathbf{w}^{T}\mathbf{x})\}^{2} -\lambda (\mathbf{w}^{T}\mathbf{w}-1) \\
\end{aligned}
\end{equation}
the gradient $\nabla\mathcal{L}(\mathbf{w},\lambda)$ becomes
\begin{equation}
\label{eq:grad2}
\nabla\mathcal{L}(\mathbf{w},\lambda) =\sum_{k=1}^{p} \mathrm{E}\{G_{k}(\mathbf{w}^{T}\mathbf{x})\}\mathrm{E}\{G_{k}^{'}(\mathbf{w}^{T}\mathbf{x})\mathbf{x}\} -2\lambda\mathbf{w} =\mathbf{0}
\end{equation}
where $\lambda = \frac{1}{2}\sum_{k=1}^{p}\mathrm{E}\{G_{k}(\mathbf{w}^{T}\mathbf{x})\}\mathrm{E}\{G_{k}^{'}(\mathbf{w}^{T}\mathbf{x})\mathbf{w}^{T}\mathbf{x}\}$, and the structure of Hessian matrix $\nabla^{2}\mathcal{L}(\mathbf{w},\lambda)$  becomes complex 
\begin{equation}
\label{eq:hessian2}
\begin{aligned}
\nabla^{2}\mathcal{L}(\mathbf{w},\lambda) &=\sum_{k=1}^{p}\mathrm{E}\{G_{k}^{'}(\mathbf{w}^{T}\mathbf{x})\mathbf{x}\}\mathrm{E}\{G_{k}^{'}(\mathbf{w}^{T}\mathbf{x})\mathbf{x}^{T}\}\\
& + \sum_{k=1}^{p} \mathrm{E}\{G_{k}(\mathbf{w}^{T}\mathbf{x})\}\mathrm{E}\{G_{k}^{''}(\mathbf{w}^{T}\mathbf{x})\mathbf{x}\mathbf{x}^{T}\} -2\lambda \mathbf{I}\\
\end{aligned}
\end{equation}
When more nonlinear functions are required to improve the negentropy estimation ($p>1$), the inversion of Hessian matrix cannot be simplified compared with the diagonal approximation in equation (\ref{eq:hessian}), and the efficient fixed-point method in $FastICA$ fails. 

\section{Proposed method}
\label{section:3}
Given the expectations of linearly independent functions $\overline{\mathbf{G}}(y_{i})=(\overline{G}_{1}(y_{i}),\cdots,\overline{G}_{p}(y_{i}))^{T}$ , minimum discrimination information\cite{gokhale1978} is aimed at determining the distribution $p_{i}(y_{i})$, which is closest to the prior $p_{0}(y_{i})$ in KL divergence.
\begin{equation}
\label{eq:mdi}
\begin{aligned}
\min_{p_{i}} \quad & KL(p_{i}||p_{0})=\int p_{i}(y_{i})\log\frac{p_{i}(y_{i})}{p_{0}(y_{i})} dy_{i}\\
\mathrm{s.t.} \quad & \int p_{i}(y_{i})G_{k}(y_{i}) dy_{i}=c_{k} \quad k=1,2,\ldots,p \,\\
\quad \quad & \int p_{i}(y_{i})dy_{i}=1
\end{aligned}
\end{equation}
The prior $p_{0}(y_{i})$ used in ICA is the standard Gaussian $\phi(y_{i})$, and the solution to the above optimization is Gibbs distribution with prior,
\begin{equation}
\label{eq:gibbs}
p_{i}(y_{i}) =\frac{\phi(y_{i})e^{f_{i}(y_{i})}}{\int \phi(y_{i})e^{f_{i}(y_{i})}dy_{i}}
\end{equation}
where $f_{i}(y_{i})=\mathbf{\beta}_{i}^{T}\overline{\mathbf{G}}(y_{i})$, $\mathbf{\beta}_{i}$ contains the coefficients concerning nonlinear functions. The form of $p_{i}(y_{i})$ in equation (\ref{eq:gibbs}) is similar to the exponentially tilted Gaussian (used in $ProDenICA$\cite{HaTi2002}), and $\int \phi(y_{i})e^{f_{i}(y_{i})}dy_{i}$ is the partition function. We then substitute the $p_{i}(y_{i})$ in $KL(p_{i}||\phi)$ to obtain the minimum discrimination information $KL^{min}(p_{i}||\phi)$,
\begin{equation}
\label{eq:mdi2}
\begin{aligned}
&KL^{min}(p_{i}||\phi)\quad\\
& =\int \frac{\phi(y_{i})e^{f_{i}(y_{i})}}{\int \phi(x)e^{f_{i}(y_{i})}dy_{i}} \log\frac{\frac{\phi(y_{i})e^{f_{i}(y_{i})}}{\int \phi(y_{i})e^{f_{i}(y_{i})}dy_{i}}}{\phi(y_{i})} dy_{i}\\
%&=\frac{\int\phi(y_{i})e^{f_{i}(y_{i})}f(y_{i})dy_{i}}{\int \phi(y_{i})e^{f_{i}(y_{i})}dy_{i}}-\log \int %\phi(y_{i})e^{f_{i}(y_{i})}dy_{i}\\
&=\int\phi(y_{i})e^{f_{i}(y_{i})}f_{i}(y_{i})dy_{i} -\int \phi(y_{i})e^{f_{i}(y_{i})}dy_{i} +1
\end{aligned}
\end{equation}
The last expression in equation (\ref{eq:mdi2}) can be easily proved by noticing that $KL^{min}(p_{i}||\phi)$'s invariance of the scale of partition function $\int \phi(y_{i})e^{f_{i}(y_{i})}dy_{i}$ and the maximal value (of the last expression) is obtained when $\int \phi(y_{i})e^{f_{i}(y_{i})}dy_{i}=1$. We will maximize $KL^{min}(p_{i}||\phi)$ as the contrast function in our ICA method. In the $MDIICA$, several assumptions are likely to be true:
\begin{enumerate}
	\item Since $KL(p_{i}||\phi)$ works as the measure concerning the departure from standard Gaussian, $KL^{min}(p_{i}||\phi)$ is a lower-bound for KL divergence and the maximization of $KL^{min}(p_{i}||\phi)$ may 
	lead to the maximization of true $KL(p_{i}||\phi)$;
	\item Owing to the definition of minimum discrimination information, it is reasonable to deem that the unknown $p_{i}(y_{i})$ in $KL^{min}(p_{i}||\phi)$ is close to the standard Gaussian $\phi(y_{i})$.
\end{enumerate}
Although the two assumptions are partly similar to the maximum entropy used in $FastICA$\cite{Hyv1998,Aapo2001}, we will show their main differences in the rest of the section.

Maximizing the total minimum discrimination information $\sum_{i=1}^{m}KL^{min}(p_{i}||\phi)$ can be viewed as a joint maximization over the unmixing matrix $\mathbf{W}$ and the density distributions of sources' estimation $\mathbf{y}$, fixing one argument and maximizing over the other. The optimization problem in our ICA algorithm is
\begin{equation}
\label{eq:mdi3}
\begin{aligned}
\max_{\{\mathbf{w}_{i},p_{i}\}_{i=1}^{m}} \quad & \sum_{i=1}^{m}KL^{min}(p_{i}||\phi)\quad\mathrm{s.t.}  &  \mathbf{W}^{T}\mathbf{W}=\mathbf{I}
\end{aligned}
\end{equation}
where $\mathbf{w}_{i}$ is the $i_{th}$ row in the unmixing matrix $\mathbf{W}$. The joint maximization in equation (\ref{eq:mdi3}) consists of two iterative stages:
\begin{itemize}
	\item $\max_{\{p_{i}\}_{i=1}^{m}}\sum_{i=1}^{m}KL^{min}(p_{i}||\phi)$. Fixing $\mathbf{W}$, each $p_{i}$ is estimated by minimizing the weighted least squares concerning the approximation of MDI.
	\item $\max_{\{\mathbf{w}_{i}\}_{i=1}^{m}}\sum_{i=1}^{m}KL^{min}(p_{i}||\phi)$. Given $p_{i}$, $\mathbf{W}$ is restricted to be orthonormal and is calculated via the fixed-point method\cite{fastica1999,HaTi2002}.
\end{itemize} 
Similar joint maximization has been used in past researches concerning projection pursuit \cite{PPDE1984,EPP1987} and ICA \cite{HaTi2002,Sam2012}.
\subsection{Second-order approximation of minimum discrimination information}
To simplify the integral in equation (\ref{eq:mdi2}), we construct a grid of $L$ (500) values $y_{i}^{*l}$ with $\Delta$ step, and let the corresponding frequency $q_{i}^{*l}$ be
\begin{equation}
\label{eq:freq}
q_{i}^{*l} = \sum_{j=1}^{N}\mathbb{I}(y_{i}^{j}\in \left(y_{i}^{*l}-\Delta/2,y_{i}^{*l}+\Delta/2\right])/N
\end{equation}
where $\mathbb{I}(.)$ is the indicator function, and $y_{i}^{j}=\mathbf{w}_{i}^{T}\mathbf{x}_{j}\,(j=1,\cdots,N)$.
Thus, the original $KL^{min}(p_{i}||\phi)$ in equation (\ref{eq:mdi2}) is converted to the following form
\begin{equation}
\label{eq:mdi4}
\begin{aligned}
KL^{min}(p_{i}||\phi) &=\sum_{l=1}^{L}\{q_{i}^{*l}f_{i}(y_{i}^{*l}) - \Delta\phi(y_{i}^{*l})e^{f_{i}(y_{i}^{*l})}\} +1\\
\end{aligned}
\end{equation}
This is similar to the generative additive models\cite{gam1990} used in $ProDenICA$\cite{HaTi2002}, which can be solved by a sequence of iterative reweighted least squares (IRLS)\cite{wolk1988}. However, we don't adopt that strategy in our method, we utilize the definition of minimum discrimination information to cut down the computation burden instead. Since $p_{i}(y_{i})$ is close to the standard Gaussian $\phi(y_{i})$ and the partition function $\int \phi(y_{i})e^{f_{i}(y_{i})}dy_{i}$ is equal to 1 at the maximal point of $KL^{min}(p_{i}||\phi)$ , we can conclude that $f_{i}(y_{i})$ is close to the zero. The second-order approximation of $p_{i}(y_{i})$ is
\begin{equation}
\label{eq:approx}
p_{i}(y_{i}) = \phi(y_{i})e^{f_{i}(y_{i})} \approx \phi(y_{i})(1+f_{i}(y_{i})+\frac{1}{2}f_{i}^{2}(y_{i}))
\end{equation}
Substituting equation (\ref{eq:approx}) into equation (\ref{eq:mdi4}), the original maximization of $KL^{min}(p_{i}||\phi)$ becomes equivalent to the minimization of weighted least squares below
\begin{equation}
\label{eq:wls}
\begin{aligned}
\min_{f_{i}} \quad &\sum_{l=1}^{L} \Delta\phi(y_{i}^{*l})\left(f_{i}(y_{i}^{*l})-\frac{q_{i}^{*l}-\Delta\phi(y_{i}^{*l})}{\Delta\phi(y_{i}^{*l})}\right)^{2}\\
\end{aligned}
\end{equation}
Two nonlinear functions $\overline{\mathbf{G}}(y_{i})=(\overline{G}_{1}(y_{i}),\overline{G}_{2}(y_{i}))^{T}$ have been used in projection pursuit \cite{cook1993} and negentropy estimation\cite{Hyv1998}, and they are appropriate to be the basis functions in most cases.
\begin{equation}
\label{eq:G1}
\overline{G}_{1}(y_{i}) = y_{i}e^{-\frac{y_{i}^{2}}{2}}\quad \overline{G}_{2}(y_{i}) = e^{-\frac{y_{i}^{2}}{2}}
\end{equation}
Since the size of the nonlinear basis used in equation (\ref{eq:wls}) is constant and we can solve the weighted least squares efficiently in linear time. Compared with $ProDenICA$ and $FastICA$, our method has successfully replaced a sequence of iterative reweighted least squares (in $ProDenICA$)\cite{HaTi2002} with a single weighted least squares, and the complex constraints in equation (\ref{eq:constraints1})(\ref{eq:constraints2}) are avoided.

%\begin{algorithm}[!t]  
%	\caption{Maximizing the approximated minimum discrimination information }  
%	\label{alg:wls}  
%	\begin{algorithmic}[1]
%		\FOR{$i = 1$ to $m$}
%		\STATE{$y_{i}=\mathbf{w}_{i}^{T}\mathbf{x}$}
%		\STATE{construct a grid of $L$ values $y_{i}^{*l}$ with $\Delta$ step, and calculate the frequency $q_{i}^{*l}$ in equation (\ref{eq:freq}) }
%		\STATE{$\overline{\mathbf{G}} =\left(\overline{\mathbf{G}}(y_{i}^{*1}),\cdots,\overline{\mathbf{G}}(y_{i}^{*L})\right)^{T}$}
%		\STATE{$\mathbf{\Phi}=diag(\Delta\phi(y_{i}^{*1}),\cdots,\Delta\phi(y_{i}^{*L}))$}
%		\STATE{$\mathbf{z}=\left(\frac{q_{i}^{*1}-\Delta\phi(y_{i}^{*1})}{\Delta\phi(y_{i}^{*1})},\cdots,\frac{q_{i}^{*L}-\Delta\phi(y_{i}^{*L})}{\Delta\phi(y_{i}^{*L})}\right)$}
%		\STATE{solve weighted least squares in equation (\ref{eq:wls})}
%		\STATE{$\beta_{i}=\left(\overline{\mathbf{G}}^{T}\mathbf{\Phi}\overline{\mathbf{G}}\right)^{-1}\overline{\mathbf{G}}^{T}\mathbf{\Phi}\mathbf{z}$}
%		\STATE{$f_{i}(y_{i})=\beta_{i}^{T}\overline{\mathbf{G}}(y_{i})$}
%		\ENDFOR
%	\end{algorithmic}  
%\end{algorithm} 
\subsection{Fixed-point method}
Given the fixed $p_{i}(y_{i})$, the partition function $\int \phi(y_{i})e^{f_{i}(y_{i})}dy_{i}=1$ and the minimum discrimination information in equation (\ref{eq:mdi2}) develops into the following form
\begin{equation}
\label{eq:mdi5}
\begin{aligned}
KL^{min}(p_{i}||\phi)\quad
=\int\phi(y_{i})e^{f_{i}(y_{i})}f_{i}(y_{i})dy_{i}
=\mathrm{E}\{f_{i}(\mathbf{w}_{i}^{T}\mathbf{x})\}
\end{aligned}
\end{equation}
Different from  $FastICA$, we can apply the fixed-point method to $KL^{min}(p_{i}||\phi)$ directly no matter how many nonlinear functions are used. The additive model (\ref{eq:mdi5}) is easier to be optimized compared with the sum of quadratic terms in equation (\ref{eq:neg}).
\begin{algorithm}[h]  
	\caption{Estimating the \emph{unmixing matrix} $\mathbf{W}$ by fixed-point method}  
	\label{alg:fixed-point}  
	\begin{algorithmic}[1]
		\FOR {$i = 1$ to $m$}             
		\STATE {$\mathbf{w}_{i} \gets \mathrm{E}\{\mathbf{x}f^{'}_{i}(\mathbf{w}_{i}^{T}\mathbf{x})\}-\mathrm{E}\{f^{''}_{i}(\mathbf{w}_{i}^{T}\mathbf{x})\}\mathbf{w}_{i}$}
		\ENDFOR
		\STATE{$\mathbf{W}\gets(\mathbf{w}_{1},\cdots,\mathbf{w}_{m})^{T}$}
		\STATE{symmetric decorrelation}
		\STATE{$\mathbf{W}\gets(\mathbf{W}\mathbf{W}^{T})^{-\frac{1}{2}}\mathbf{W}$}
	\end{algorithmic}  
\end{algorithm}
\section{Experiments and results}
\label{section:4}
\subsection{Implementation details}
\begin{figure}[h]
	%\vskip 0.2in
	\centering
	\centerline{\includegraphics[width=0.4\linewidth]{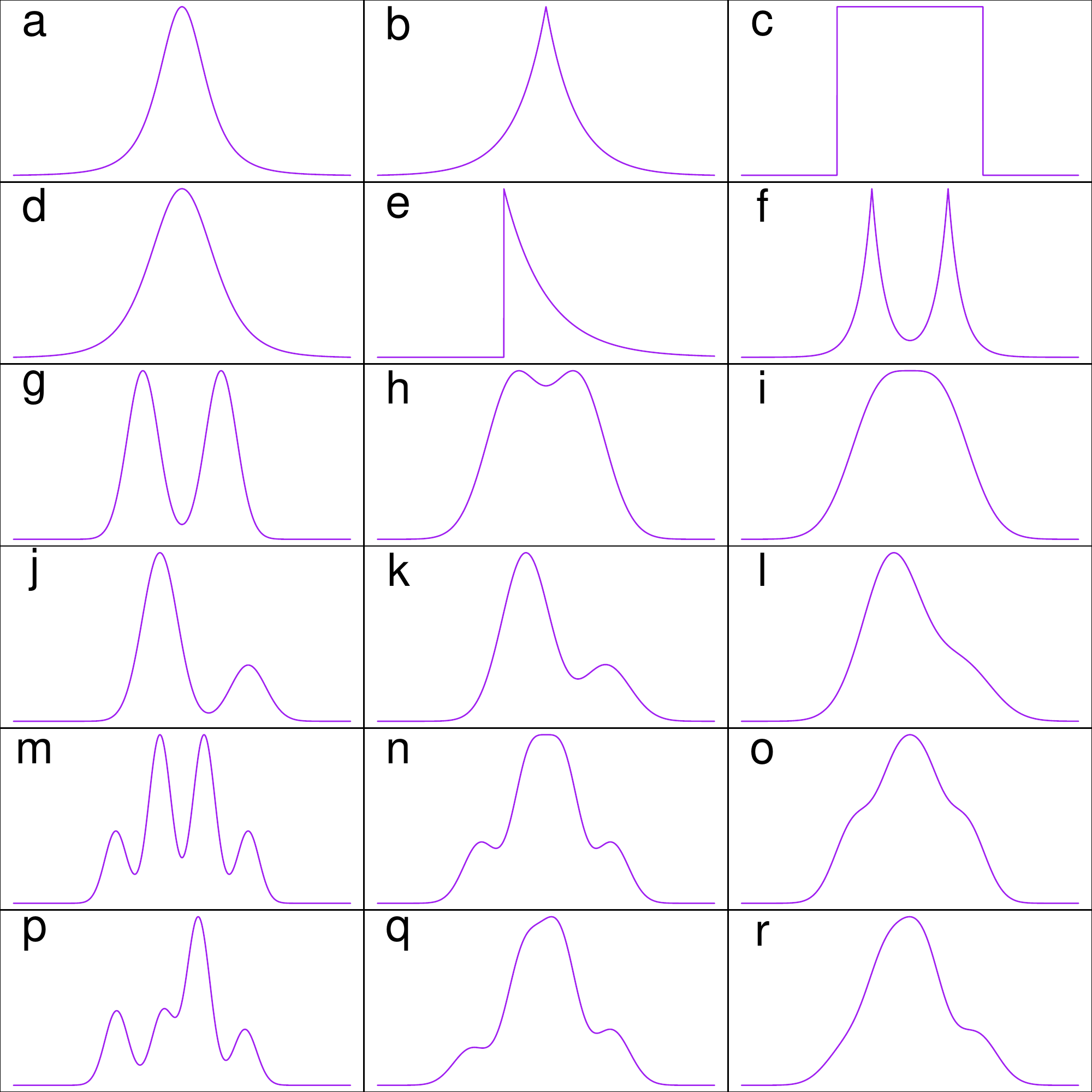}}
	\caption{Eighteen probability density functions of sources.}
	\label{18distributions}
	%\vskip -0.2in
\end{figure}

Two experiments are implemented to test the performance of the proposed method. The first experiment has been conducted in the past researches \cite{Bach2003,RADICAL,HaTi2002}, where the independent sources' components $s_{i}$ are chosen from 18 distributions\cite{Bach2003} in Figure \ref{18distributions}. The second experiment is designed to validate our method with real signals \cite{nordhausen2008tools}.

%\begin{figure}[!ht]
%	\vskip 0.2in
%	\begin{center}
%		\centerline{\includegraphics[width=1.5in]{18Distributions.pdf}}
%		\caption{Eighteen probability density functions of sources.}
%		\label{18distributions}
%	\end{center}
%	\vskip -0.2in
%\end{figure}

%\begin{figure*}[!t]
%	\centering
%	\subfloat[]{\includegraphics[width=1.5in]{./figs/18Distributions.pdf}
%		\label{18distributions}}
%	\hfil
%	\subfloat[]{\includegraphics[width=1.5in]{./figs/2dim.pdf}
%		\label{2dimN1000}}
%	\caption{Left : Eighteen probability density functions of sources. Right : Average Amari metrics (multiplied by 100) for two-component ICA. The overall mean Amari metrics and CPU Elapsed time (ms) is recorded in the legend. }
%	\label{heart}
%\end{figure*}

\begin{figure}[h]
	%\vskip 0.2in
	\centering
	\centerline{\includegraphics[width=0.8\linewidth]{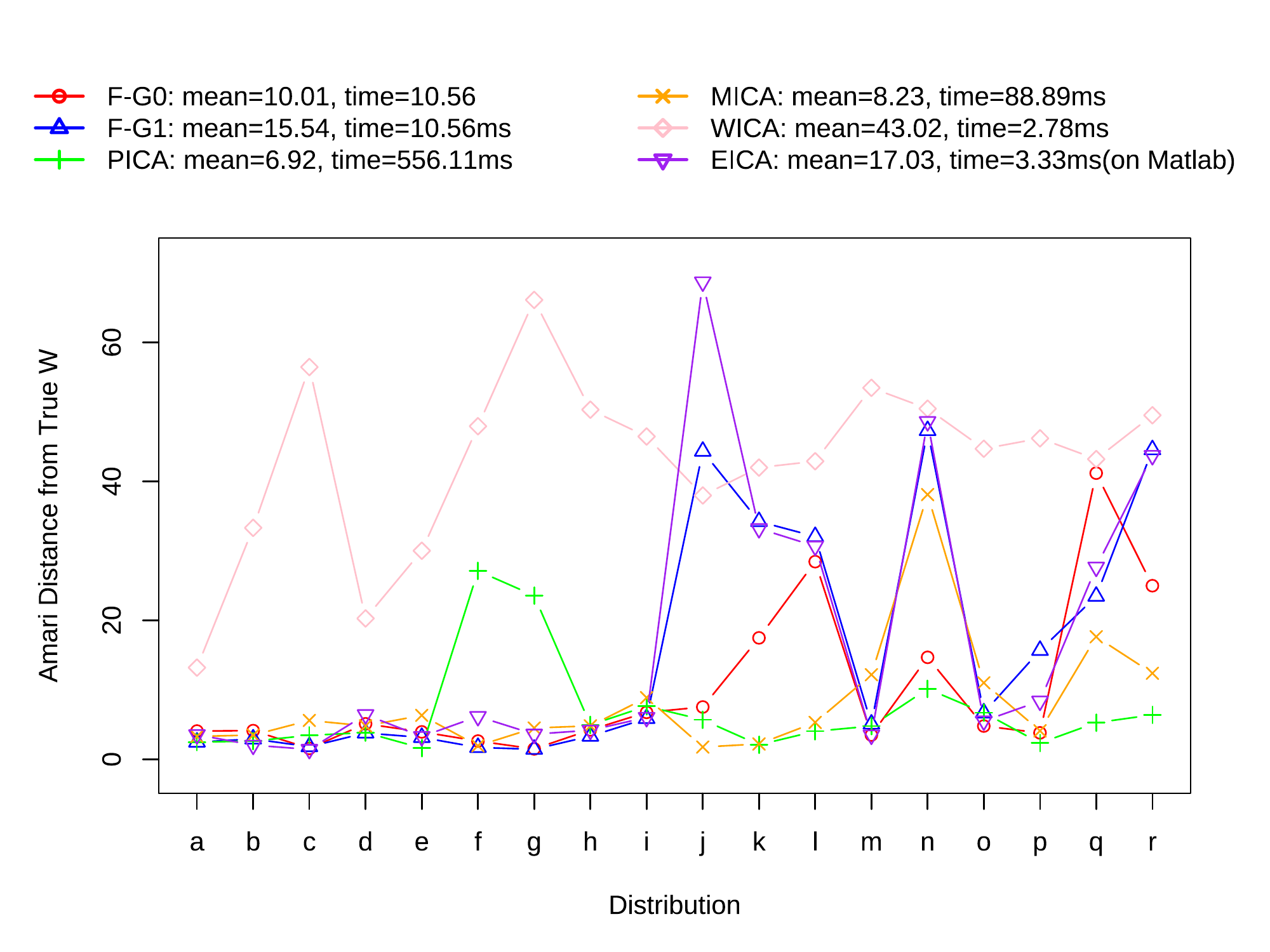}}
	\caption{Average Amari metrics (multiplied by 100) for two-component ICA. The overall mean Amari metrics and CPU Elapsed time (ms) is recorded in the legend.}
	\label{2dimN1000}
	%\vskip -0.2in
\end{figure}
Several existing algorithms are chosen for comparisons, the implementation details are presented in Table \ref{table:methods}. $EFICA$ \cite{Kol2006} is a statistically efficient version of the FastICA, and $WeICA$ \cite{spurek2018} is the recent ICA algorithm based on the weighted second moments. It might not be appropriate to compare the CPU elapsed time between $EFICA$ (Matlab implementation) and other ICA algorithms (R implementation), thus we denote the EFICA's CPU elapsed time with superscript $^{*}$.
\begin{table}[]
	\caption{ICA methods used in the experiments.}
	\label{table:methods}
	\centering
	\begin{tabular}{llll} 
		\toprule
		Methods&Symbols  &Parameters  &Sources   \\ 
		\midrule
		$FastICA$\cite{fastica1999}&F-G0  &$G0=\frac{y_{i}^{4}}{4}$ &$\emph{ProDenICA}$\quad package\cite{proDenICApackage}   \\
		$FastICA$&F-G1  &$G1=\log cosh(y_{i})$  &$\emph{ProDenICA}$\quad package   \\
		$ProDenICA$\cite{HaTi2002}&PICA  &Gfunc=GPois   & $\emph{ProDenICA}$\quad package \\
		$MDIICA$&MICA  & equation (\ref{eq:G1})  &   \\
		$EFICA$\cite{Kol2006} &EICA& default& \emph{EFICA} \quad package \cite{EFICA}\\
		$WeICA$\cite{spurek2018} &WICA& / & our own implementation  \\
		\bottomrule
	\end{tabular}
\end{table}
The separation performance of ICA is measured by the value of Amari metrics $d(\mathbf{W},\mathbf{W}_{0})$\cite{Amari1999}, which is equal to zero if and only $\mathbf{W}$ and $\mathbf{W}_{0}$ are equivalent,
\begin{equation}
\label{Amarimetric}
\begin{aligned}
d(\mathbf{W},\mathbf{W}_{0}) &=\frac{1}{2m}\sum_{i=1}^{m}\left(\frac{\sum_{j=1}^{m}|r_{ij}|}{\max_{j}|r_{ij}|}-1\right)\\
&+\frac{1}{2m}\sum_{j=1}^{m}\left(\frac{\sum_{i=1}^{m}|r_{ij}|}{\max_{i}|r_{ij}|}-1\right)
\end{aligned}
\end{equation}
where $r_{ij}=(\mathbf{W}\mathbf{W}_{0}^{-1})_{ij}$, $\mathbf{W}_{0}$ is the known truth. 

%\begin{wrapfigure}{l}{0.5\textwidth}
%	\vskip 0.2in
%	\begin{center}
%		\centerline{\includegraphics[width=0.55\linewidth]{./figs/18Distributions.pdf}}
%		\caption{Probability density functions of sources: $(a)$ Student with
%			3 degrees of freedom; $(b)$ double exponential; $(c)$ uniform; $(d)$ Student with
%			5 degrees of freedom; $(e)$ exponential; $(f)$ mixture of two double exponentials;
%			$(g)-(h)-(i)$ symmetric mixtures of two Gaussians: multimodal, transitional and
%			unimodal; $(j)-(k)-(l)$ nonsymmetric mixtures of two Gaussians, multimodal, tran-
%			sitional and unimodal; $(m)-(n)-(o)$ symmetric mixtures of four Gaussians: mul-
%			timodal, transitional and unimodal; $(p)-(q)-(r)$ nonsymmetric mixtures of four
%			Gaussians: multimodal, transitional and unimodal.}
%		\label{18distributions}
%	\end{center}
%	\vskip -0.2in
%\end{wrapfigure}

\subsection{Experiments with simulated signals}
For each distribution in Figure \ref{18distributions}, a pair of independent components ($N=1000$) is generated as sources in two-dimensional ICA, then they are mixed by a random invertible matrix to produce the mixtures $\mathbf{x}$. This experiment is replicated 100 times for each distribution, the average Amari metrics and CPU elapsed time are recorded in Figure \ref{2dimN1000}. $FastICA$ works well only when its single nonlinear function is close to the unknown sources' distribution, otherwise its separation performance or negentropy estimation might degenerate due to the unwanted density mismatching. As can be seen in Figure \ref{2dimN1000}, F-G0 and F-G1 perform well in symmetric distributions (f, g, h, i) thanks to their symmetric nonlinear functions (F-G0: $G0=\frac{y_{i}^{4}}{4}$, F-G1: $G1=\log cosh(y_{i})$) used in negentropy estimation, whereas their separation performance deteriorates in the cases of nonsymmetric or nontrivial distributions (j, k, l, p, q, r, n). Compared with the single nonlinear function used in F-G0 and F-G1, the nonparametric estimation used in PICA and nonlinear functions used in MICA are more flexible in these nontrivial cases. The CPU elapsed time required by MICA (88.89 ms) is much less than PICA's (556.11 ms), as a result of transforming a sequence of IRLS (in PICA) into a single weighted least squares (in MICA).
%\begin{figure}[!ht]
%	\vskip 0.2in
%	\begin{center}
%		\centerline{\includegraphics[width=3.5in]{./figs/2dim.pdf}}
%		\caption{Average Amari metrics (multiplied by 100) for two-component ICA. The overall mean Amari metrics and CPU Elapsed time (ms) is calculated in the legend. Average Amari metrics of MICA is second lowest and the corresponding CPU elapsed time is acceptable compared with the PICA.}
%		\label{2dimN1000}
%	\end{center}
%	\vskip -0.2in
%\end{figure}

\subsection{Experiment with real data}
We design an image separation experiment in this subsection, where the three gray-scale images (depicting a forest road, cat, and sheep) used are from the \emph{ICS} package \cite{nordhausen2008tools}. We vectorize them to arrive into a $130^{2}\times 3$ data matrix, then mix them by a random invertible matrix (100 replications). The results are recorded in Table \ref{table:ICS}. As can be seen in Table \ref{table:ICS}, F-G0, F-G1, MICA, EICA suffer from the large Amari metrics, whereas PICA acquires the best separation performance with the longest CPU elapsed time. To improve the separation performance of MICA, we introduce another two nonlinear functions ($G0, G1$ in Table \ref{table:methods}) to the negentropy estimation, then an efficient version MICA$_{4}$ is immediately available.
The average Amari metrics of MICA$_{4}$ is the second lowest and the corresponding CPU elapsed time is $44\%$ of PICA's.

\begin{table}[]
	\caption{ICA experiments on ICS images (100 replications).}
	\label{table:ICS}
	%\centering
\begin{tabular}{llllllll}
	\toprule
	Mean& F-G0  & F-G1  & PICA & MICA & WICA & MICA$_{4}$&EICA  \\
	\midrule
	Amari metrics& 40.79 & 55.12 & 19.4& 48.83& 30& 28.96&42.55\\
	Elapsed time(ms) & 225.26 & 134.66& 1683.41 & 401.16& 1226.42&741.37   & 29.50$^{*}$\\
	\bottomrule
	\\
	\toprule
	Standard deviation& F-G0  & F-G1  & PICA & MICA & WICA & MICA$_{4}$&EICA  \\
	\midrule
	Amari metrics& 7.36 & 3.17 & 3.57& 10.13& 0.0& 0.0&5.07\\
	Elapsed time(ms) & 7.52 & 4.62& 37.78 & 51.22& 14.03&67.50   & 18.88$^{*}$\\
	\bottomrule
\end{tabular}
\end{table}

\section{Conclusion}
In this paper, we propose a novel ICA algorithm based on the second-order approximation of minimum discrimination information. Our algorithm alleviates the difficulties in $FastICA$ when more nonlinear functions are required in negentropy estimation. In addition, we reduce the computation burden by transforming a sequence of IRLS in $ProDenICA$\cite{HaTi2002} into a single weighted least squares. The proposed method is concise and efficient, several experiments validate its performance compared with other ICA methods. 
\label{section:5}

% \usepackage{multirow}

% or


%\section*{Acknowledgment}


\ifCLASSOPTIONcaptionsoff
  \newpage
\fi

\bibliographystyle{IEEEtran}
\bibliography{ref}

\end{document}